\documentclass[runningheads]{llncs}

\usepackage[T1]{fontenc}

\usepackage{graphicx}
\usepackage{soul}
\usepackage{xcolor}
\usepackage{hyperref}
\usepackage{navigator}

\begin{document}

\title{Chess Rating Estimation from Moves and Clock Times Using a CNN-LSTM}

\author{Michael Omori \orcidID{0009-0000-4632-9272}, Prasad Tadepalli \orcidID{0000-0003-2736-3912}}

\authorrunning{M. Omori and P. Tadepalli}

\institute{Oregon State University, Corvallis OR, USA \\
\email{\{omorim, prasad.tadepalli\}@oregonstate.edu}}

\maketitle              

\begin{abstract}

Current chess rating systems update ratings incrementally and may not always accurately reflect a player's true strength at all times, especially for rapidly improving players or very rusty players. To overcome this, we explore a method to estimate player ratings directly from game moves and clock times. We compiled a benchmark dataset from Lichess with over one million games, encompassing various time controls and including move sequences and clock times. Our model architecture comprises a CNN to learn positional features, which are then integrated with clock-time data into a Bidirectional LSTM, predicting player ratings after each move. The model achieved an MAE of 182 rating points on the test data. Additionally, we applied our model to the 2024 IEEE Big Data Cup Chess Puzzle Difficulty Competition dataset, predicted puzzle ratings and achieved competitive results. This model is the first to use no hand-crafted features to estimate chess ratings and also the first to output a rating prediction after each move. Our method highlights the potential of using move-based rating estimation for enhancing rating systems and potentially other applications such as cheating detection.

\end{abstract}

\outline{1}{1 Introduction}
\section{Introduction}

Traditional rating systems like Elo \cite{elo1978rating} and Glicko \cite{glickman2012example}, though effective, often require many games to accurately reflect the true level of skill of a player, leading to potential mismatches and inaccuracies, especially for new accounts or rapidly improving players. This delay in accurate rating adjustment can skew competitive fairness and hinder effective matchmaking. For example, lower-rated players were shown to have longer winning streaks than higher-rated players on average, suggesting that their rating may be less accurate compared to higher-rated players \cite{chowdhary2023quantifying}. This could occur if a person creates a new account and is either an experienced chess player who made a second private account or a new player who transferred skills from other board games. In both cases, their initial rating starts lower on sites such as lichess \cite{lichess} and chess.com \cite{chesscom}, but their performance could be much higher.

To overcome these limitations, we propose a novel method that estimates player ratings directly from game moves and clock times, aiming to provide accurate ratings after each game. Our dataset comes from Lichess, encompasses 1.2 million games from April 2021 to July 2024 (inclusive), covers various time controls and includes detailed move sequences and clock times. Chess streamers like GothamChess \cite{gothamchess} have attempted this on their own (without computer aid) by using their chess knowledge to look at games and "guess the elo" and it is quite challenging to be consistently accurate.

Our method is a neural network architecture that combines a four-layer Convolutional Neural Network (CNN) and a Bidirectional Long Short-Term Memory (LSTM) network. The CNN extracts positional features from the chessboard, which are then integrated with clock time data by the LSTM to predict player ratings move-by-move. This approach allows the model to consider both spatial and temporal dimensions of the game, leading to more precise rating predictions. The model achieved a mean absolute error (MAE) of 183 rating points in the test data, demonstrating the effectiveness of the overall approach.

Ablation studies confirm that the inclusion of clock time data improves the model performance. Additionally, the same model is tested on the 2024 IEEE Big Data Cup Chess Puzzle Difficulty Competition dataset, where it predicts puzzle ratings competitively, indicating the model's versatility and robustness.

An additional application of the rating estimate for each move is to check for anomalous behavior. For example, a 1300 rated player that is playing like a 2600 in half their games, but 1300 in the other half may be considered an anomaly. A CNN was \cite{patria2021cheat} trained on Lichess games but only achieved marginally better results than random guessing for classifying cheaters. Lichess itself has software called Kaladin for detecting cheaters \cite{lichesskaladin}. It uses a CNN and also utilizes chess insights, which includes metrics such as average centipawn loss and number of moves. One downside to this approach is that it can only do as well as the people who labeled the data.

Our main contribution is creating the first neural network that predicts chess rating move by move without hand-crafted features. This is unlike the rating players normally get after playing an entire game, which is based off their current rating, the game result, and the opponent's rating. Second, we collated a benchmark from Lichess data for evaluation and finally demonstrated the utility of the amount of clock time remaining after each move in estimating player rating.

In conclusion, this method presents an advancement in chess rating estimation by leveraging deep learning techniques to analyze in-game moves and clock times. The promising results and potential applications highlight the importance of this method, paving the way for more dynamic and fair rating systems in competitive chess.

\begin{credits}
\subsubsection{\discintname}
The authors have no competing interests to declare that are
relevant to the content of this article.
\end{credits}

\outline{1}{2 Background}
\section{Background}
There are various rating systems to estimate a player's skill, such as FIDE's Elo and Lichess's Glicko 2. Elo uses expected scores and a K-factor for rating updates, while Glicko incorporates rating deviation (RD) and volatility for more accurate adjustments. We will briefly explain Glicko2 since that is the rating system that we are trying to estimate through moves.

\subsection{Glicko-2 Rating System}

Glicko-2 incorporates volatility to indicate how consistent a player plays, along with a rating and a rating deviation for each player. A collection of games like in a tournament is used to update these values and is treated to have occurred simultaneously with the same ratings $\mu$, volatility, and deviation $\phi$ for each game. The rating deviation is manually set for each player to a specified amount such as between 0.3 and 1.2.

First, the variance v is calculated based on the outcomes of the game for each opponent with the rating $\mu_{j}$.

\begin{equation}
    v = \left[ \sum_{j=1}^{m} g(\phi_j)^2 E(\mu, \mu_j, \phi_j) \left\{ 1 - {E}(\mu, \mu_j, \phi_j) \right\} \right]^{-1}
\end{equation}
where
\begin{equation}
    g(\phi) = \frac{1}{\sqrt{1 + 3 \phi^2/\pi^2}}
\end{equation}

\begin{equation}
    {E}(\mu, \mu_j, \phi_j) = \frac{1}{1 + \exp(-g(\phi_j)(\mu - \mu_j))}
\end{equation}

The change in rating $\Delta$ is then calculated by comparing the actual outcome of the game (win 1, draw 0.5, loss 0) with the expected result based on the logistic formula in equation 3. Note that $g(\phi_{J})$ and v are directly proportional to the step size of the rating change.

\begin{equation}
    \Delta = v \sum_{j=1}^{m} g(\phi_j) \left( s_j - {E}(\mu, \mu_j, \phi_j) \right)
\end{equation}

The new volatility is calculated using the Illinois algorithm \cite{snyder1953inverse}, based on $\Delta$, $\phi$, and $v$. A more detailed explanation can be found in the paper by Mark Glickman \cite{glickman2012example}. The final step involves updating both the rating deviation and the player ratings to $\mu'$ and $\phi'$.

\begin{equation}
    \phi^* = \sqrt{\phi^2 + \sigma'^2}
\end{equation}

\begin{equation}
    \phi' = \frac{1}{\sqrt{\frac{1}{\phi^{*2}} + \frac{1}{v}}}
\end{equation}

\begin{equation}
    \mu' = \mu + \phi'^{2} \sum_{j=1}^{m} g(\phi_j) \left( s_j - {E}(\mu, \mu_j, \phi_j) \right)
\end{equation}

Overall, the Glicko-2 rating update is applied after each game, while our work tackles the challenge of rating updates after each move within a game.

\outline{1}{3 Related Works}
\section{Related Works}

One of the first works on modeling chess player skill used Rybka 3 \cite{rybka} evaluations to model player moves based off rating \cite{regan2011intrinsic}. Later, 30 hand-crafted features were fed into random forests \cite{breiman2001random} and support vector machines \cite{cortes1995support} to predict ratings on players binned by rating into 10 different splits \cite{tijhuis2023predicting}. Rather than focusing the problem as a classification problem with limited bins, we choose to frame it as a regression problem, as this gives us more flexibility and precision in rating players. Also, the data from the previous rating estimation paper is not publicly available, so we compiled our own open source dataset from Lichess for this task.

The recent popularity of word prediction has permeated its way into the chess field as move prediction. Maia \cite{mcilroy2020aligning} adapted AlphaZero's residual Convolutional Neural Network (CNN) \cite{silver2018general}, removed the search algorithm to predict moves based on rating bins, and improved the accuracy over a depth-limited Stockfish. Stockfish is currently the strongest chess engine \cite{stockfish} and the original Leela Zero was an open source clone of it, although it has since migrated to a transformer architecture \cite{monroe2024mastering}. The Maia model was then fine-tuned to predict specific player moves \cite{mcilroy2022learning}. GPT-2 was fine-tuned on 2.8 million chess games, demonstrating strategic understanding without directly obtaining a rating. Another transformer model achieved a 2895 Lichess Blitz rating through supervised learning on Stockfish evaluations from 10 million games \cite{ruoss2024grandmaster}.

Further classification of chess games has been explored through a binary classifier for detecting cheaters, using the Euclidean distance to compare player features such as move accuracy and response time \cite{laarhoven2022towards}. One paper showed that the winning likelihood strongly correlates with the remaining time in rapid chess games, essentially less time lowers chances of winning \cite{sigman2010response}. Multi-class classification of chess games, predicting outcomes such as win/draw/loss based on moves and ratings with an LSTM, has also been attempted \cite{reddy2023chess}. Another study predicted the brilliance of moves (a binary classification) using features from chess engine analysis \cite{zaidi2024predicting}. While these approaches are similar to ours, they focus on classification tasks, whereas our model predicts a continuous value.

Rating prediction has been conducted in other gaming domains, such as Massive Online Battle Arena (MOBA) games. In one case, a transformer was trained to predict player ratings for matchmaking using descriptive features like kills and deaths recorded in three-minute intervals \cite{zhang2022quickskill}. The predicted rating showed a higher correlation with the win rate than the existing rating system, particularly for players with fewer than 15 games.

\outline{1}{4 Dataset}
\section{Dataset}
1.2 million Lichess games from April 2021 to July 2024 are used in the dataset, with 30,000 games from each month. \footnote{Games are taken after March 2021 because of some incorrect results that month in the Lichess database due to a datacenter fire} Lichess is a chess website chosen because it is open source and contains the largest publicly available database of games for free use.  The data is split randomly with 80\% used for the training set and 20\% used for the test set. Other available input features include the amount of clock time remaining after each move. The outputs to be estimated are the ratings of both players for a single game.

Moreover, we applied the same model architecture to the 2024 IEEE Big Data 2024 Cup Chess Puzzle Difficulty Competition \cite{ieee_bigdata2024_chess_puzzle}. This data contains a Forsyth-Edwards Notation (FEN) \cite{fen} string representation of the starting position, along with the moves in Universal Chess Interface (UCI) \cite{uci} notation for the solution.

\subsection{Time Controls}
Lichess time controls are less than 29 seconds for UltraBullet, 179 for Bullet,  479 for Blitz, 1499 for Rapid, else Classical. Games with increment add 40 times the increment to the total time \cite{lichesstime}. A time control of "5+3", which is 5 minutes per side with an additional 3 seconds for each move, becomes 5 x 60 + 40 x 3 = 420 seconds. 420 seconds is less than 479 and is considered blitz. 40 comes from an approximation of the average number of moves in a game.

\subsection{Metrics}

The Mean Absolute Error (MAE) is used to provide an intuitive metric to understand the performance of models. The MAE is given for the pre-standardized rating to allow for a more interpretable loss. The test data is split up by time control to see how model performance varies between time controls. The metric used for the Chess Puzzle Competition is the mean squared error (MSE). We also provide the MSE for the game ratings.

\outline{1}{5 Method}
\section{Method}
\subsection{Input Features}
We use a similar input representation as AlphaZero, with one plane for each piece-color combination, totaling 12 planes \cite{silver2018general}. Each square contains a 1 if the corresponding piece is present; otherwise, it includes a 0. For example, the plane for black pawns contains 1s in the seventh row and 0s everywhere else. Following \cite{sigman2010response}, the remaining clock-time is included as a feature. That paper showed a strong correlation between winning likelihood, remaining time, and position evaluation in 3-minute chess games. The remaining clock-time feature is standardized by subtracting the mean and dividing by the standard deviation. The same standardization process is applied to the ratings for more stable learning. The mean rating is 1514, with a standard deviation of 366. The same input is used for the puzzle competition, except it does not have clock time features because they were unavailable. FEN string representations were converted to the array representation.

\subsection{Model Architecture}
The architecture, which we will refer to as RatingNet, is a CNN that takes in each board state and passes this into a Bidirectional Long Short Term Memory Neural Network (LSTM) to predict both player ratings for the game after each move. CNNs have demonstrated success in various computer vision tasks, having been initially applied to document recognition \cite{lecun1998gradient}. LSTMs have also been proven to be successful in modeling long-term dependencies \cite{hochreiter1997long}. A four-layer CNN is used with batch normalization and mean pooling after each layer. The CNN is kept shallow to mitigate overfitting without having to train on a lot of data. The output of the CNN is concatenated with the amount of clock time spent for that corresponding move. This is then fed through the LSTM, which has two fully connected layers on top to output the ratings for the players of that game. The LSTM takes in the hidden state representation at the previous move and clock usage as input. One dropout layer is used after the first fully connected linear layer. Leaky ReLU activations are employed \cite{xu2020reluplex}. A simple baseline for predicting the mean chess rating is also shown for comparison.

RatingNet does not use one player's rating as input to predict the other player's rating. The reason is to focus the network on understanding chess rather than just predicting a rating close to the other player's rating. This is to mitigate potential problems when the rating difference is significant, one player's rating is provisional, or when a player is playing on an alternate account.

\begin{figure}[ht]
    \centering
    \includegraphics[width=1\textwidth]{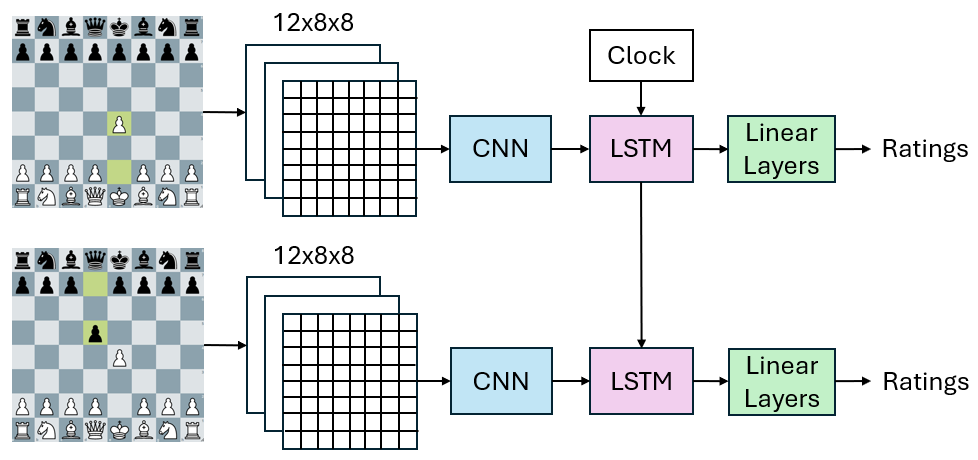}
    \caption{The model architecture used to predict chess ratings after each move using a CNN and LSTM taking in the remaining clock time feature. This shows the first two time steps of a sample game.}
    \label{fig: arch}
\end{figure}

\outline{1}{6 Results}
\section{Results}
\begin{table}[ht]
\centering
\caption{MAE of Different Methods Across Various Chess Time Controls}\label{tab:chess_loss_comparison}
\begin{tabular}{|r|r|r|r|}
\hline
\textbf{Time Control} & \textbf{RatingNet} & \textbf{RatingNetNoClock} & \textbf{Mean} \\
\hline
Average Test Loss & 182 & 239 & 346 \\
Ultrabullet Loss & 186 & 269 & 250 \\
Bullet Loss & 182 & 274 & 340 \\
Blitz Loss & 183 & 244 & 378 \\
Rapid Loss & 182 & 207 & 305 \\
Classical Loss & 151 & 185 & 246 \\
\hline
\end{tabular}
\end{table}

\begin{table}[ht]
\centering
\caption{MSE of Different Methods Across Various Chess Time Controls}\label{tab:chess_mseloss_comparison}
\begin{tabular}{|r|r|r|r|}
\hline
\textbf{Time Control} & \textbf{RatingNet} & \textbf{RatingNetNoClock} & \textbf{Mean} \\
\hline
Average Test Loss & 56,618 & 91,619 & 183,538 \\
Ultrabullet Loss & 56,927 & 115,833 & 95,459 \\
Bullet Loss & 57,569 & 117,776 & 176,296 \\
Blitz Loss & 57,330 & 94,048 & 216,803 \\
Rapid Loss & 56,433 & 71,215 & 141,016 \\
Classical Loss & 38,491 & 56,032 & 94,304 \\
\hline
\end{tabular}
\end{table}

The average MAE across the time controls is 182 for RatingNet. This is reasonable given that some chess categories, such as the United States Chess Federation (USCF), are 200 points per class \cite{uscf}. The loss for longer-time controls is lower. The most likely reason for this is that the range of ratings decreases as the time control increases, partially because faster time controls are more popular. In addition, faster time controls may have a bit more randomness because players have less time to think, making it harder to estimate the skill levels.

The results may need to be taken with a grain of salt however because the percentage of times that the white player won with a higher rating and when the black player won with a higher rating was calculated only to be 53\% in the training data. Such rating problems also appear in over-the-board chess games and possible factors for this center on the fact that rating systems are often overoptimistic for players with higher ratings due to the simplicity of their model \cite{glickman1999rating}. Another consideration is that there is no rating label for each move, but only for the player after having played a certain number of games. The rating outputs after each move is modeling the rating estimate of the player after having seen that move and all the previous moves made by the players in that game.

We additionally report the MSE on the test set of the IEEE Big Data 2024 Cup Chess Puzzle Difficulty Competition. On the public leaderboard, the MSE was 82,049. The score is comparable to the MSE for RatingNetNoClock on estimating rating in games.

\outline{1}{7 Clock Ablation}
\section{Clock Ablation}
Because faster time controls have less time for thinking and are closer to random play than longer time controls, our hypothesis was that the correlation between move quality and chess rating may weaken, thus having additional information such as the amount of time spent on each move could be useful. To test this, an ablation experiment was performed to determine the usefulness of the clock time feature. Without the clock time feature (setting the clock input to a constant stream of 0s), the model does slightly worse; specifically, the addition of clock information reduces the MAE by 57 points, a 24\% improvement. Notably, the clock feature helps significantly at faster speeds, such as bullets, with a 92-point reduction in error, or a 34\% improvement. When the time control is longer, for example, in classical, the error is the lowest, with a 151 MAE for RatingNet. Overall, the network is able to utilize the time spent per move as an effective feature for skill estimation.

\outline{1}{8 Game Analysis}
\section{Game Analysis}
A couple of sample games are investigated to determine how different types of moves affect the rating estimation of the model. We sample a game shown in Fig. 2 in which the model performs well. It correctly lowered the rating estimate of the player of the white pieces after the move Nf5, which allows the knight to be taken by the opponent's bishop. The full game is shown in the appendix as game 1. We also analyze a game with high loss, shown in the appendix as game 2. This bullet game was played between a white rating of 2922 and a black rating of 3163. The final predicted rating for white is 1459 and for black it is 1460. Because the average rating is around 1500, we do expect higher ratings to present more difficulties because there are fewer examples of such high rated games. We expect that in some games however, RatingNet would provide more accurate rating estimations than the player's actual rating as players do have off days or games where they play really well. With an accurate enough rating estimation system, one could consider further extending this work to potentially detect engine use or external assistance. For example, if a specific move could be marked as having a much higher rating than what the player's true strength is, further investigation could occur.

\begin{figure}[ht]
    \centering
    \includegraphics[width=0.5\textwidth]{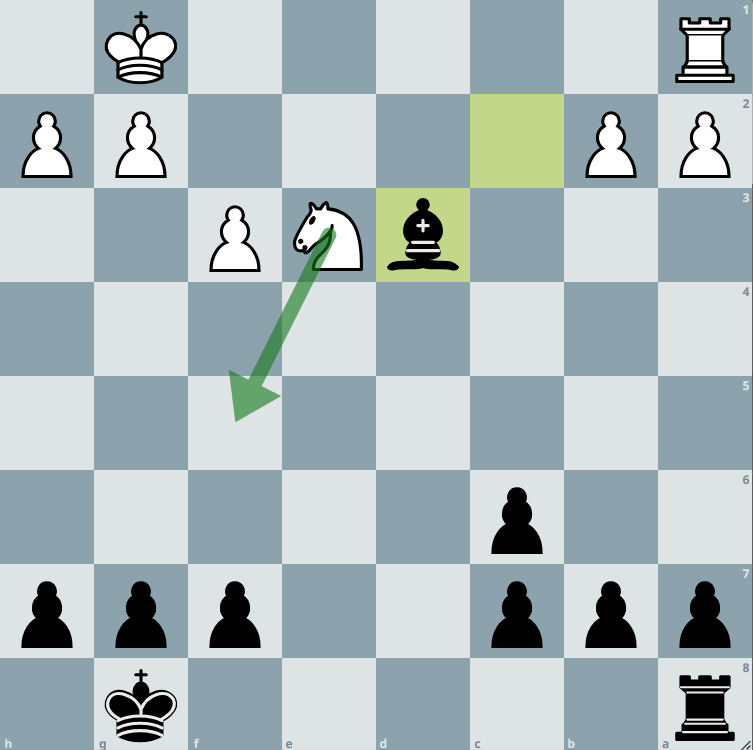}
    \caption{White plays Nf5, a blunder because black can take it with the bishop on d3. White's estimated rating goes from 1255 to 1241. Their actual Lichess rating in this bullet game is 1224.}
    \label{fig:example}
\end{figure}

\outline{1}{9 Conclusion}
\section{Conclusion}
Our proposed method for online chess rating estimation leverages game moves and clock times without relying on hand-crafted features for a variety of time controls. By integrating a CNN-LSTM architecture, we demonstrated the potential to accurately predict player rating on a move-by-move basis. The performance of the model, with an MAE of 182, shows promise in improving rating systems, providing real-time and dynamic assessments of the player's skills. This novel approach could also serve as a foundation for applications beyond rating estimation, such as anomaly detection in games and paves the way for more granular rating systems in chess

\outline{1}{10 Appendix}
\section{Appendix}
\subsection{Hyperparameters and code}

training batch size: 32, learning rate: 1e-4, weight decay: 1e-5, epochs: 50, learning rate patience: 10, learning rate factor: 0.5, dropout: 0.5.

The model was trained on one A40 gpu and took 12 hours to finish training. The code: \href{https://github.com/AstroBoy1/RatingNet}{https://github.com/AstroBoy1/RatingNet}

\subsection{Game 1}
1. e4 e5 2. Nf3 Nc6 3. d4 exd4 4. Nxd4 Nf6 5. Nxc6 dxc6 6. Nc3 Qxd1
7. Nxd1 Bc5 8. Bd3 O-O 9. O-O Re8 10. Be3 Bxe3 11. Nxe3 Nxe4
12. Bxe4 Rxe4 13. Nd1 Bf5 14. f3 Re2 15. Rf2 Rxc2 16. Rxc2 Bxc2
17. Ne3 Bd3 18. Nf5 Bxf5 19. g4 Bg6 20. Kf2 Rd8 21. Kg3 Rd2
22. Rb1 Bxb1 23. h4 Rxb2 24. h5 Rxa2 25. g5 Bf5 26. f4 h6 27. Kh4 hxg5
28. Kxg5 Bd3 29. h6 gxh6 30. Kxh6 Rg2 31. f5 f6 32. Kh5 Be2 33. Kh6 Rg5 1/2-1/2

\subsection{Game 2}
1. e4 g6 2. d4 Bg7 3. Nc3 d6 4. h4 Nf6 5. f3 Nc6 6. Be3 h5 7. Nge2 e5 8. d5 Ne7 9. Qd2 O-O 10. O-O-O a6 11. Bh6 b5 12. Bxg7 Kxg7 13. g4 hxg4 14. fxg4 Bxg4 15. a3 Rb8 16. Bg2 b4 17. axb4 Rxb4 18. Qd3 Qb8 19. b3 a5 20. Kd2 Qb6 21. Qe3 Bxe2 22. Qxb6 cxb6 23. Kxe2 Rc8 24. Kd3 Rd4 25. Ke2 Rxc3 0-1

\bibliographystyle{splncs04}
\bibliography{bibliography}

\outline{1}{References}

\end{document}